\documentclass{article}
\usepackage{hyperref}
\usepackage{mlspconf}
\usepackage{graphicx}
\usepackage{amsthm}
\usepackage{amsmath}
\usepackage{amssymb}
\usepackage{color}
\usepackage{parskip}
\usepackage{mwe}
\usepackage[utf8]{inputenc}
\usepackage[inline]{enumitem}

\usepackage{multirow}
\usepackage{adjustbox}
\usepackage{bm}
\usepackage{tikz}
\usepackage{pgfplots}
\usepackage{pgfplotstable}
\usetikzlibrary{arrows}
\pgfplotsset{compat=1.14}
\usepgfplotslibrary{statistics}

\newlist{inlinelist}{enumerate*}{1}
\setlist*[inlinelist,1]{%
  label=(\roman*),
}
\newcommand\blfootnote[1]{%
  \begingroup
  \renewcommand\thefootnote{}\footnote{#1}%
  \addtocounter{footnote}{-1}%
  \endgroup
}

\title{Graph topology inference benchmarks for machine learning}

\name{{Carlos Lassance$^{\diamond}$, Vincent Gripon$^{\diamond}$, and Gonzalo Mateos$^{\dagger}$}
\thanks{This work was supported in part by the Brittany region and by the NSF award CCF-1750428.}}

\address{$^{\diamond}$ IMT Atlantique, Lab-STICC, France, $^{\dagger}$ Dept. of ECE, University of Rochester, USA}

\begin{document}

\maketitle
\begin{abstract}
Graphs are nowadays ubiquitous in the fields of signal processing and machine learning. As a tool used to express relationships between objects, graphs can be deployed to various ends: \begin{inlinelist} \item clustering of vertices, \item  semi-supervised classification of vertices, \item supervised classification of graph signals, and \item denoising of graph signals.\end{inlinelist} However, in many practical cases graphs are not explicitly available and must therefore be inferred from data. Validation is a challenging endeavor that naturally depends on the downstream task for which the graph is learnt. Accordingly, it has often been  difficult to compare the efficacy of different algorithms. In this work, we introduce several ease-to-use and publicly released benchmarks specifically designed to reveal the relative merits and limitations of graph inference methods. We also contrast some of the most prominent techniques in the literature.
\end{abstract}
\begin{keywords}
Graph learning, network topology inference, benchmarks, graph signal processing, machine learning.
\end{keywords}

\section{Introduction}
\blfootnote{978-1-7281-6662-9/20/\$31.00 ©2020 IEEE}
Graphs are mathematical objects that express relationships between items, referred to as vertices. As a natural representation of complex data structure, graphs are ubiquitous, in particular in the field of machine learning, where they can be used for various ends: \begin{inlinelist}\item they can model the inner dependencies of observations, e.g. functional connectivity in the brain~\cite{GSPFunctional}; \item they can model the relationship between data samples, e.g. social networks and citation graphs~\cite{sen2008collective}; and \item they can be used to directly model data, e.g. gene-expression levels collected from microarray experiments~\cite{subramanian2005gene}.\end{inlinelist}

However, graphs are not always explicitly available. Many recent works have therefore considered the problem of inferring the topology (i.e. the edges) of the graph based on nodal observations~\cite{pasdeloup2017characterization,kalofolias2018large,shekkizhar2019graph}; see also~\cite{gonzalo_spmag_19} for a recent tutorial treatment. Inferring a graph structure can be performed in a task-agnostic manner, where only unsupervised observations are considered. In such a case, priors are used to relate observations to the sought graph structure: e.g. smoothness~\cite{kalofolias2018large}, stationarity~\cite{pasdeloup2017characterization}, sparsity~\cite{shekkizhar2019graph}, and probabilistic~\cite{egilmez2017jstsp} as well as graph filtering-based~\cite{shafipour2018topoidnsTSP18} generative models, just to name a few. Other works consider inferring a graph for a specific task. For example in~\cite{koopman2016information} the authors infer graphs for medical search. In~\cite{iscen2019label,lassance2019improved,hu2020exploiting}, the authors aim at improving the accuracy of various classification tasks using inferred graphs: semi-supervised learning, visual based localization and few-shot learning.

Because it is more general, task-agnostic graph inference is of particular interest. In this context, common desiderata are to generate graphs used for visualization~\cite{kalofolias2018large} and interpretation~\cite{anirudh2017influential}. On the other hand, it is challenging to compare methods for which there is no ground truth. Unsurprisingly, many works rely on synthetic data to evaluate the ability of their proposed methods in unveiling the topology from the observations. While synthetic data are always useful to perform controlled scalability experiments as well as reveal the emerging statistical and computational trade-offs, this validation protocol comes with two shortcomings. First, the models used to generate synthetic data are likely to be biased in favor of the proposed methods. Second, the ability of the proposed method to handle hard real-world problems is often not demonstrated convincingly.

In order to address this problem, standardized benchmarks are required. The main challenge is that benchmarks are necessarily task-specific, and as such they do not encompass the whole potential offered by state-of-the-art methods. To fill in this gap, in this work we introduce a broad collection of benchmarks that are specifically designed to reveal the relative merits and limitations of graph inference algorithms. To this end, we consider three timely problems arising with network data: \begin{inlinelist}\item unsupervised clustering of vertices; \item semi-supervised classification of vertices (with or without vertex features); and \item graph signal denoising.\end{inlinelist} For each problem we introduce a balanced and easy-to-use dataset that we release publicly
\footnote{\scriptsize{\url{https://github.com/cadurosar/benchmark_graphinference}}}. Note that our work is meant to benchmark the graph inference task, for a benchmark of the unsupervised/semi-supervised methods themselves we refer the reader to OGN~\cite{hu2020ogb}. 
Furthermore, the released datasets comprise various types of signals, namely natural images, audio, texts, and traffic information. Note that we do not include brain data and protein-protein interactions that are two of the most interesting use-cases of graph inference and classification. Our choice is informed by recent developments in the literature~\cite{he2020deep,Errica2020A}, that have found no significant performance gains when graph-based machine learning techniques are brought to bear for some tasks in these areas. As our objective here is to compare graph inference methods and not the techniques used in the downstream tasks, we will not delve into this issue further. 

All in all, the contributions of this work can be summarized as follows. We assemble a diverse set of datasets for various tasks and types of signals, meant to assess the efficacy of graph inference methods. We compare selected prominent methods from the literature and identify their relative strengths and shortcomings across different tasks and types of data. We provide a public release of prepackaged data and a simple script to evaluate future methods, facilitating comparisons with the graph learning algorithms considered in this paper. The outline for the remainder of this paper is as follows. In Section~\ref{statement} we introduce the considered tasks. In Section~\ref{datasets}, we present the created datasets. In Section~\ref{methods}, we review some of the main methods from the literature. In Section~\ref{experiments}, we perform experiments and discuss the results. Finally, we conclude in Section~\ref{conclusion}.

\section{Problem statement}
\label{statement}

We consider three tasks that can benefit from graph inference methods. These tasks were chosen to represent most widely considered applications cases found in the recent literature.

Before getting to the details of the methods, let us make a quick recall about graphs and graph signals. A graph $G = \langle V, \mathbf{W} \rangle$ is a tuple where $V$ is the finite set of vertices and $\mathbf{W} \in \mathbb{R}^{|V|\times |V|}$ is the adjacency matrix. Typically, $\mathbf{W}$ is symmetric. The degree matrix of the graph is the diagonal matrix $\mathbf{D}_\mathbf{W}$ where $\mathbf{D}_\mathbf{W}[i,i] = \sum_j{\mathbf{W}[i,j]}$. Note that $\mathbf{W}[i,j]$ refers to the weight located at $i$-th row and $j$-th column in $\mathbf{W}$. Some authors like to consider normalized adjacency matrices, such as $\mathbf{W} \leftarrow \mathbf{D}_{\mathbf{W}}^{-1/2} \mathbf{W} \mathbf{D}_\mathbf{W}^{-1/2}$ or $\mathbf{W} \leftarrow \mathbf{D}_{\mathbf{W}}^{-1} \mathbf{W}$. The graph Laplacian is the matrix $\mathbf{L} = \mathbf{D}_\mathbf{W} - \mathbf{W}$. A graph signal is a (most of the time real-valued) matrix $\mathbf{X}\in \mathbf{R}^{|V|\times F}$, where $F$ stands for the number of nodal features.

Because the Laplacian is symmetric and real-valued, it can be eigendecomposed as $\mathbf{L} = \mathbf{F} \bm{\Lambda} \mathbf{F}^\top$, where $\mathbf{F}$ is orthonormal and $\mathbf{F}^\top$ is its transpose; and $\bm{\Lambda}$ is diagonal, and its elements are sorted in ascending order. We refer to the first columns of $\mathbf{F}$ as the low-frequency eigenvectors of the Laplacian.

Consider a given set of observations, each one composed of several features. We divide our benchmarks into two types of machine learning tasks. In the first ones (Tasks 1 and 2), the graph model dependencies between observations. As such, a vertex in the graph corresponds to one observation, and is associated with the corresponding features. In the second one (Task 3), the graph models relationships between features. Therefore, here a vertex in the graph represents a feature and the graph is used as a proxy to the topology of the signal. We expect some 
methods will perform better on the first series of tasks and others to be more adequate to the second one.

\subsection{Task 1: Unsupervised Clustering of Vertices (UCV)}

Consider a dataset composed of $|V|=N$ observations, each one containing $F$ features. Given a number of classes $C$, we consider the task of partitioning the $N$ observations into $C$ classes, such that the variability inside classes is smaller than the variability between classes. In practice, variability can be measured using various metrics. For the purpose of obtaining quantified benchmarks, we consider here that the observations belong to $C$ categories (e.g. classes of images or sounds), and that this information is not available when processing the considered methods. So, the performance of a considered method is evaluated by computing the Adjusted Mutual Information score~\cite{vinh2009information} based on the ground truth.

Note that this clustering problem can be treated without a graph structure. Examples are using $C$-means or DB-Scan algorithms. In the context of this work, we consider using spectral clustering. Spectral clustering consists in creating a graph linking the observations where the edges are inferred from the corresponding features. Then, vertices are projected using the first eigenvectors of the graph Laplacian and clustered using standard non-graph methods. In our work, we use the discretization method first proposed in~\cite{yu2003clustering} when features have been projected onto the first $C$ eigenvectors of the graph Laplacian except the very first one. We use the default SciKit-Learn~\cite{scikit-learn} implementation of spectral clustering and of the $C$-means algorithm in our experiments. 

\subsection{Task 2: Semi-Supervised Classification of Vertices (SSCV)}

Consider a dataset composed of $|V|=N$ observations, each one containing $F$ features. Here, a portion of the $N$ observations are labeled. The task consists in inferring the labels of the other portion of observations. Again, we consider datasets where we have access to the ground truth, and artificially hide the labels of part of the observations when processing the data. The score consists in measuring the accuracy of the classification on initially unlabeled observations.

This problem can be solved without relying on graphs. For example, a common solution would consist in performing a supervised classification using only the labeled observations. In this work, we consider inferring a graph connecting observations from the features. Then, we use this graph in two settings. In the first setting, we want the graph to fully encompass the information contained in the features, and therefore perform label propagation. Label propagation 
consists in diffusing the labels from the known observations to the other ones using the inferred graph structure. In a second setting, we use both the graph structure and the features to perform classification. We use the methodology described in~\cite{wu2019simplifying}, called Simplified Graph Convolution (SGC), where the goal is to combine feature diffusion with logistic regression. Note that in our case we should obtain equivalent results for both SGC and GCN, as the number of observed nodes is smaller than the amount of features as noted by~\cite{vignac2020choice}.

In more detail, we use two layers of feature diffusion ($\hat{\mathbf{X}} = \mathbf{W}^2\mathbf{X}$), followed by a logistic regression. The models are trained for 100 epochs, using Adam optimization with a learning rate of $0.001$. We use the average over 100 runs of the accuracy using random splits of 5\% training set and 95\% test set. We always report the average accuracy and standard deviation. To propagate labels, we simply diffuse the label signal one time using the exponential of the adjacency matrix. We note that SGC models tend to use the ``normalized augmented adjancency matrix'' $\tilde{\mathbf{W}} = \mathcal{\mathbf{I}} + \mathbf{W}$ where $\mathcal{\mathbf{I}}$ is the identity matrix. This augmented adjacency matrix is then normalized $\tilde{\mathbf{W}} \leftarrow \mathbf{D}_{\tilde{\mathbf{W}}}^{-1/2} \tilde{\mathbf{W}} \mathbf{D}_{\tilde{\mathbf{W}}}^{-1/2}$. In our work we test both the adjacency matrix and the augmented adjacency matrix and their respective normalizations and we report the best possible combination in terms of mean accuracy.

\begin{table*}[ht]
 \begin{center}
\caption{Summary of the tested graph topology inference methods.}
\label{table:summarytests}

\begin{tabular}{|c|c|c|c|c|}
\hline
Method     & Similarity/Distance               & $k$                                        & $\sigma$       & Adjacency matrices     \\ \hline
Naive      & \multirow{2}{*}{Cosine, Covariance, RBF}           & \multirow{3}{*}{\parbox{2.9cm}{5, 10, 20, 30, 40, 50, 100, 200, 500, 1000}} & None                  & \multirow{3}{*}{\parbox{3.1cm}{$\mathbf{W}$, $\mathbf{D}_{\mathbf{W}}^{-1/2} \mathbf{W} \mathbf{D}_\mathbf{W}^{-1/2}$, $\tilde{\mathbf{W}}$, $\mathbf{D}_{\tilde{\mathbf{W}}}^{-1/2} \tilde{\mathbf{W}} \mathbf{D}_{\tilde{\mathbf{W}}}^{-1/2}$}} \\ \cline{1-1} \cline{4-4}
NNK~\cite{shekkizhar2019graph}        &  &                                                    & \multirow{2}{*}{$10^{-4}$} &                           \\ \cline{1-2}
Kalofolias~\cite{kalofolias2018large} & Square Euclidean distance         &                                                    &                       &                           \\ \hline
\end{tabular}

\end{center}
\vspace{-0.5cm}
\end{table*}

\subsection{Task 3: Denoising of Graph Signals (DGS)}

Consider a dataset comprising $N$ observations, each one consisting of $|V|=F$ features. Consider some additive noise generated according to a distribution $\mathcal{N}$. The task consists in recovering initial observations from their noisy versions. We measure performance by looking at the Mean Squared Error (MSE) between both.

Here, the graph connects features of observations. The idea is to use the graph structure to easily segregate components of the noise from components of the initial signals. In our work, we 
use a Simoncelli low-pass filter on the graph to perform denoising. 
In our experiments we use the PyGSP~\cite{pygsp} implementation of the Simoncelli filter, which is defined by its spectral response as follows:

$$f_{l}=\begin{cases} 1 & \mbox{if }\lambda_l\leq \frac{\tau}{2}\\
            \cos\left(\frac{\pi}{2}\frac{\log\left(\lambda_l\right)}{\log(2)}\right) & \mbox{if }\frac{\tau}{2}<\lambda_l\leq\tau\\
            0 & \mbox{if }\lambda_l>\tau \end{cases},$$

where $\tau\in [0,1]$ is a user-defined threshold and $\lambda_l$ the $l$-th Laplacian eigenvalue. We normalize the eigenvalues by dividing by the largest one, so that $0\leq \lambda_l \leq 1$. We vary the parameter $\tau$ from 0 to 1 in increments of 0.025. We use the noisy signal with a SNR (Signal to Noise Ratio) of 7, from~\cite{irion2016efficient}, and report the best SNR found for each graph construction.




\section{Datasets}
\label{datasets}

For Tasks 1 and 2, we use datasets of images, audio and texts (documents). To reduce the difficulty of the tasks in the image and audio domains, we choose to use features extracted from pretrained deep neural networks. Task 3 (DGS) data comes from real life traffic information. Additional details are given in the coming paragraphs.

\subsection{Image dataset}

For the image dataset we use the training set portion of the ``102 Category Flower Dataset'' (shortened as flowers102)~\cite{Nilsback08}. This split contains $N=1020$ images of $C=102$ classes of flowers (10 images per class). The features are extracted from the final pooling layer of the Inceptionv3 architecture~\cite{szegedy2016rethinking}, which has a size of $F=2048$ dimensions. Note that Inceptionv3 was trained on the 2012 split of ImageNet challenge, so that the features we obtain are a case of transfer learning. This should be one of the most challenging scenarios we consider, as it provides the highest number of classes and has the highest signal dimension to number of items ratio: 2. 

\subsection{Audio dataset}

For audio data, we use ``ESC-50: Dataset for Environmental Sound Classification''~\cite{piczak2015dataset}. This dataset contains $C=50$ classes, with 40 audio signals each (2000 in total). It also contains 5 standard splits that are not used here (as we do unsupervised and semi-supervised classification). We use the feature extractor introduced in~\cite{kumar2018knowledge} to generate our dataset, that was trained on AudioSet. Similar to the images data, this can be considered as transfer. At the end we have $N=2000$ items with $F=1024$ dimensions each. The signal dimension to number of items ratio is 0.512.

\subsection{Text dataset}

We use the cora dataset~\cite{sen2008collective}, which is composed of $N=2708$ scientific articles of $C=7$ different domains for document clustering or classification. The features come from a word indicator vector (i.e. Bag of Words BoW) that indicates if one of the words in the dictionary ($F=1433$ in total) is present on the title or abstract of the document. We prefer simple BoW because our first tests using features extracted from pretrained networks led to worse performance. The dictionary is built with the most common words in the dataset. The signal dimension to number of items ratio is: 0.53. Note that this dataset is classically used for graph semi-supervised learning as it comes with a citation graph. But in our work we completely disregard this graph. Comparisons between the ground truth graph and inferred ones could be an interesting addition to this work. But since the citation graph is not exactly redundant with the signals, it is expected that inferred graphs and citation ones are quite different. 

\subsection{Toronto traffic data denoising (Toronto)}
We use data from the road network of the city of Toronto, from~\cite{irion2016efficient}. 
 It describes traffic volume data 
 at intersections in the road network of Toronto for a total of $F=2202$ vertices and $N=1$ observation. Note that extra information is available, such as the position of each road and intersection, but our baselines only consider the raw signal data.

\section{Graph inference methods}
\label{methods}

In our work, we perform experiments using off-the-shelf graph inference techniques from the literature. We also provide implementations of the chosen techniques. Table~\ref{table:summarytests} presents a summary of the methods and variations we tested.

\vspace{-0.4cm}

\subsection{Naive baselines}

We first consider naive baselines 
by combining three steps:
\begin{enumerate}
    \item Choosing a similarity measure to be applied to either features of each vertex for Tasks 1 and 2 or to observations for Task~3. In more details, we consider cosine similarity, sampled covariance or an RBF kernel applied on the $L_2$ distance between considered items. 
    \item Choosing a number of neighbors to be kept for each vertex. We simply use a $k$-nearest neighbor selection. Note that we symmetrize the resulting graph, so that each vertex has at least $k$ neighbors.
    \item Normalizing the obtained graph adjacency matrix.
\end{enumerate}

Note that we obtain a large number of possible combinations, and perform experiments for each one. In Section \ref{experiments} we only display the results obtained by the best combination. 

\vspace{-0.4cm}

\subsection{Sparsity-based method}

We now consider a more recent sparsity-based method. We choose NNK (Non Negative Kernel regression)~\cite{shekkizhar2019graph}, due to its simplicity and its demonstrated results on semi-supervised learning tasks. This method can be interpreted as producing representations with orthogonal approximation errors, which in turn favors sparser representations. It has two parameters: $k$, the maximum degree for each vertex, and $\sigma$ the minimum value for an edge weight (threshold). In this work we test multiple values of $k$ and fix $\sigma=10^{-4}$~\cite{kalofolias2018large}. In our experiments we use the authors implementation from~\url{https://github.com/STAC-USC/PyNNK_graph_construction}.

\vspace{-0.4cm}

\subsection{Smoothness-based method}

For our smoothness based topology inference method, we rely on a state-of-the-art approach in~\cite{kalofolias2018large}. It consists in a framework that infers the graph from an underlying set of smooth signals. As it was the case with the sparsity based method, it has two parameters: $k$ the desired mean sparsity and $\sigma$ the minimum value for an edge weight. We test the same values for these two parameters as we did for the previous method and keep the best combination. In our experiments we use the implementation from the GSP toolbox~\cite{perraudin2014gspbox}.

\section{Baseline results}
\label{experiments}

\subsection{Task 1}

For Task 1: UCV, we display both the results obtained with the inferred graph structures and with a 
$C$-means baseline.  The results are presented in Table~\ref{table:unsup}. We can see that both naive and NNK get the most consistent results, with Kalofolias having difficulties with the cora dataset.  

\begin{table}[ht]
 \begin{center}
 \vspace{-0.2cm}
\caption{Results for Task 1. Here we present the best AMI score for each inference method.}
\label{table:unsup}
  \begin{adjustbox}{max width=\columnwidth}
\begin{tabular}{|c|c|c|c|c|}
\hline
\multicolumn{1}{|l|}{Method}         & \multicolumn{1}{l|}{Inference/Dataset} & \multicolumn{1}{l|}{ESC-50} & \multicolumn{1}{l|}{cora} & \multicolumn{1}{l|}{flowers102} \\ \hline
\multicolumn{2}{|c|}{$C$-means}                                                  & 0.59                    & 0.10                    & 0.36                          \\ \hline
\multirow{3}{*}{Spectral clustering} & Naive                                     & \textbf{0.66}                      & \textbf{0.34}           & \textbf{0.45}                          \\ \cline{2-5}
                                     & NNK                                     & \textbf{0.66}             & \textbf{0.34}           & 0.44                          \\ \cline{2-5}
                                     & Kalofolias                              & 0.65                      & 0.27                    & 0.44                           \\ \hline
\end{tabular}
\end{adjustbox}
\end{center}
\end{table}

\begin{table*}[ht]
 \begin{center}
\caption{Results for Task 2. Here we present the best mean test accuracy and its standard deviation for each inference method.}
\label{table:semisup}

\begin{tabular}{|c|c|c|c|c|}
\hline
\multicolumn{1}{|l|}{Method}         & \multicolumn{1}{l|}{Inference/Dataset} & \multicolumn{1}{l|}{ESC-50} & \multicolumn{1}{l|}{cora} & \multicolumn{1}{l|}{flowers102} \\ \hline
\multicolumn{2}{|c|}{Logistic Regression}                                      & 52.92\% $\pm 1.9$              & 46.84\% $\pm 1.6$                 & 33.51\% $\pm 1.7 $                        \\ \hline
\multirow{3}{*}{Label Propagation}   & Naive                                     & 59.05\% $\pm 1.8$              & \textbf{58.86\%} $\pm 2.9$                 & 36.73\% $\pm 1.6$                         \\ \cline{2-5} 
                                     & NNK                                     & 57.44\% $\pm 2.2$              & 58.66\% $\pm 2.9$                 & 33.57\% $\pm 1.6$                         \\ \cline{2-5} 
                                     & Kalofolias                              & \textbf{59.16\%} $\pm 1.8$              & 58.60\% $\pm 3.4$                 & \textbf{37.01\%} $\pm 1.7$                      \\ \hline
\multirow{3}{*}{SGC}                 & Naive                                     & 60.48\% $\pm 2.0$            & \textbf{67.19\%} $\pm 1.5$                 & \textbf{37.73\%} $\pm 1.5$                         \\ \cline{2-5}
                                     & NNK                                     & \textbf{61.38\%} $\pm 2.0$    & 66.58\% $\pm 1.5$                 & 36.81\% $\pm 1.5$                         \\ \cline{2-5} 
                                     & Kalofolias                              & 59.36\% $\pm 2.0$              & 66.28\% $\pm 1.5$                 & 37.5\% $\pm 1.5$                         \\ \hline
\end{tabular}
\end{center}
\end{table*}

\vspace{-0.6cm}

\subsection{Task 2}

For the SSCV task, the results are presented in Table~\ref{table:semisup}. We can see that using a similarity graph as support helps when compared to a simple logistic regression. Note that this is not a 100\% fair comparison as the logistic regression is not able to exploit the unsupervised data. In this task we have two methods, Label Propagation and SGC. In the first one, Kalofolias presents the best results for both flowers102 and ESC-50, but still struggles with the cora dataset. In SGC both Kalofolias and NNK seem to not be able to improve that much over the naive baselines.

\subsection{Task 3}

For the graph signal denoising task, the results are presented in Table~\ref{table:denoising}. In this scenario we are not able to use neither cosine or covariance similarity. We compare our results with the ones we would obtain using the ground truth road map graph. Our 
RBF baselines were able to reduce the amount of noise, but not at the same level as of the real road graph. The Kalofolias smooth graph was able to achieve a better SNR than the real road graph.

\begin{table}[ht]
\vspace{-0.2cm}
 \begin{center}
\caption{Results for Task 3. Here we present the best test accuracy for each baseline.}
\label{table:denoising}
  \begin{adjustbox}{max width=\columnwidth}
\begin{tabular}{|c|c|c|c|c|}
\hline
\multirow{2}{*}{Best SNR}               & Road graph & Kalofolias & RBF NNK & RBF $k$-NN \\ \cline{2-5}
          & 10.32 & \textbf{10.41} & 9.99 & 9.80    \\ \hline
\end{tabular}
\end{adjustbox}
\end{center}
\vspace{-0.4cm}
\end{table}

\subsection{Discussion on baselines}

Over all tasks we can extract some lessons on graph inference:
\begin{enumerate}
    \item \textbf{Similarity choice:} If we have multiple non-negative realizations of the signal, cosine seems the best choice. It has competitive results on all benchmarks and it does not come with a parameter (as does RBF with $\gamma$).
    \item  \textbf{Choosing parameter $k$}: The best amount of sparsity depends not only on the dataset and task, but on the similarity that was chosen. We consider the ESC-50 dataset as an example. In the spectral clustering the best $k$ value for the $k$-NN graph was 30 for cosine, 5 for RBF and 20 for covariance. We note that in the graph denoising task, the best case was to not perform $k$-neighbors thresholding.
    \item \textbf{Normalization:} Note that only our graph denoising task does not expect a normalized graph, therefore most of our better results used normalized graphs. On the graph denoising task, normalized and non-normalized graphs had similar results.
    \item \textbf{Cora dataset}: The cora dataset is challenging not only because it is not class-balanced, but also because its features are binary (a bag of words, containing 1 if the word is present in the article and 0 if not). This could be a reason for the bad performance of both NNK and Kalofolias in this dataset.
    \item \textbf{Sparse graphs in semi-supervised problems:} In the semi-supervised tasks, the test accuracy standard deviation over the splits was very high. This could possibly be caused by the fact the sparse graphs we use here have more than one connected component, meaning that sometimes there could be sections of the graph that do not have any labeled vertices. One possible future direction would be to integrate a graph sampling algorithm to the problem in order to select which vertices we should label, instead of doing so randomly.
    \item \textbf{Naive Baselines vs. optimization approaches:} Over our tests there was no clear winner between simply doing a naive $k$-NN approach and more advanced graph topology inference techniques. Kalofolias had very good performance on the Label Propagation and Denoising tasks, while NNK was consistent in SGC and Spectral Clustering, but both were not able to consistently beat the naive baseline. On the other hand, there was a clear advantage of both Kalofolias and NNK over the naive baselines when we consider the robustness of both methods to the parameter $k$ selection.
    \end{enumerate}

\vspace{-0.4cm}

\section{Conclusion}

We have introduced graph inference benchmarks that allows us to test different graph topology inference methods in real downstream signal and information processing tasks. We have tested naive graph inference methods and more advanced techniques in the literature. This allowed us first to verify that improving the graph inference leads to better performance on the downstream tasks and to take away some guidelines for experimentation in this domain. We note that while we tested various baselines, a more thorough analysis of the results is needed. The benchmark is available online and should be easy to use to compare newer techniques. We hope that this allows for more advances in the field and we are eager to continue improving this tool as the field advances.
\label{conclusion}

\vspace{-0.2cm}

\bibliographystyle{IEEEbibvv}
\bibliography{ref}

\end{document}